\pdfoutput=1
\documentclass[11pt]{article}

\usepackage[]{emnlp2021}

\usepackage{times}
\usepackage{latexsym}

\usepackage[T1]{fontenc}
\usepackage[utf8]{inputenc}

\usepackage{microtype}

\usepackage{multirow}
\usepackage{graphicx}

\usepackage{xcolor} % already imported

\newcommand{\stitle}[1]{\vspace{3pt}\noindent{\bf #1}}

\title{Salience-Aware Event Chain Modeling for Narrative Understanding}

\author{Xiyang Zhang, Muhao Chen, \and Jonathan May \\
  Information Sciences Institute \\
  University of Southern California \\
  \texttt{\{xiyangzh, muhao, jonmay\}@isi.edu}}

\begin{document}
\maketitle
\begin{abstract}
%While event identification in text has improved, 

Storytelling, whether via fables, news reports, documentaries, or memoirs, can be thought of as the communication %of chains of interesting events
of interesting and related events that, taken together, form a concrete process. It is desirable to extract the event chains that represent such processes. %in order to reason over them. 
However, this extraction remains a challenging problem. %WE posit... 
%Natural language always evolves to communicate about the procedurally evolution of events.
%In this context, identification of useful event chain patterns that represent such evolution remains a challenging problem. 
We posit that this is due to the nature of the texts from which chains are discovered. Natural language text interleaves a narrative of concrete, salient events with background information, contextualization, opinion, and other elements that are important for a variety of necessary discourse and pragmatics acts but are not part of the principal chain of events being communicated. We introduce methods for extracting this principal chain from natural language text, by filtering away non-salient events and supportive sentences. We demonstrate the effectiveness of our methods at isolating critical event chains by comparing their effect on downstream tasks. We show that by pre-training large language models on our extracted chains, we obtain improvements in two tasks that benefit from a clear understanding of event chains: narrative prediction and event-based temporal question answering. The demonstrated improvements and ablative studies confirm that our extraction method isolates critical event chains.~\footnote{Our code and data are available at \url{https://github.com/juvezxy/Salience-Event-Chain}.}

\end{abstract}

\section{Introduction}
%The extraction of events in natural language text can be a building block for various downstream tasks. 
Human languages always communicate about evolving events. Hence, it is important for NLP systems to understand how events are procedurally described in text.
In this context, identifying patterns of event chains is important but challenging, as it requires knowledge of inter-event relations such as temporal or causal relations. Modeling high-quality event chain patterns from text would be a first step toward the more general goal of event schema induction, which involves generating high-level representations of event relations and structures. The extracted chain patterns can in turn %encode information useful
represent useful information for %popular
core natural language tasks such as question answering~\citep{reddy-etal-2019-coqa}, semantic role labeling~\citep{cheng2018implicit}, story generation~\citep{Yao_Peng_Weischedel_Knight_Zhao_Yan_2019}, and reading comprehension~\citep{ostermann-etal-2019-mcscript2}.

\begin{figure}[t]
    \centering
    \includegraphics[width=\columnwidth]{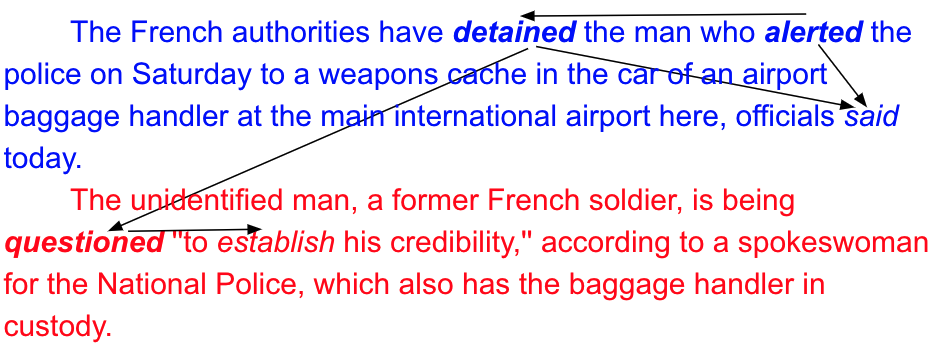}
    \caption{An illustration of event chains and salience-aware and discourse-aware filtering on an example text. The words in \textit{italic} are the events and the arrows between them show the temporal relation given by TEAR (pointing from previous to next). Salient events are shown in \textit{\textbf{bold}}; the blue sentence is in the \textit{Main Event} category whereas the red sentence is in the \textit{Evaluation} category.}
    \label{Figure: example}  
\end{figure}

Generally speaking, ``events'' correspond to what we perceive as happening around us. According to the theory of embodied cognition~\citep{wilson2002six}, our understanding of the world is shaped by various aspects of our entire bodies, involving also our language comprehension. However, it is currently difficult to gather enough data from other modalities to model real world ``events,'' and written text, especially in the news domain, seems to be our best option. 

Previous attempts have been made to generate event chains by modeling %news data
narratives in news, stories and documentaries~\citep{chambers-jurafsky-2008-unsupervised,weber-etal-2018-hierarchical,li-etal-2020-connecting}. The problem with such data is that the articles are usually a mixture of the true narrative flow with other content, which serves to explain the context or provide side information. Most prior approaches do not take into account the centrality or salience of events or the discourse structures %of news articles, 
that describe those events.
Accordingly, overlooking such important discourse properties when choosing the events of a sequence introduces noise in the modeling of event chains, and causes trivial or irrelevant content to be captured and inferred in narrative understanding tasks.

In this work, we explore the use of salience identification~\citep{liu-etal-2018-automatic-event,jindal-etal-2020-killed} and discourse profiling~\citep{choubey-etal-2020-discourse} to help isolate the main event chains from other distracting events, and show the effect on two recent datasets related to narrative understanding and temporal understanding. More specifically, we obtain event chains from documents and perform different methods of filtering, then build event language models, which we use to predict story ending events from the ROCStories dataset~\citep{mostafazadeh-etal-2016-corpus} and answer temporal event questions from the TORQUE dataset~\citep{ning-etal-2020-torque}. By comparing the use of event language models built on differently filtered event sources, we show that filtering out distracting narrative information can indeed benefit the modeling of event relations, thus leading to %potential better 
more reliable extraction of useful event chain patterns.

The main contributions in this work are three-fold. %We incorporate event salience information in a recent discourse classification model to achieve better result;
(1) To support narrative understanding tasks with high-quality event chains, we develop a new event chain extraction method that is discourse-aware, and particularly, salience-aware.  %We demonstrate through multiple experiments and analyses on two event understanding-related datasets, that salience/discourse-aware filtering of event chains can improve (a) narrative prediction, and (b) temporal commonsense understanding.
(2) We show the effectiveness of salience-based and discourse-aware filtering of event chains 
in improving narrative prediction, which leads to improvement on the Story Cloze Test. 
(3) We further demonstrate that the discourse patterns captured by the filtered event chains enhance language models on temporal understanding of events, leading to state-of-the-art performance on answering temporal ordering questions. %TORQUE~\citep{ning-etal-2020-torque}.

\section{Event Chains and Narratives}

Information extraction techniques have evolved to extract event mentions as well as their ordering information~\citep{lin-etal-2020-joint,han-etal-2019-joint,wang-etal-2020-joint}, hence enabling the automated induction of raw event chains from text. Tools such as TEAR~\citep{han-etal-2019-joint} have been developed, which we can use to extract both the events and temporal ordering from text documents, like those shown in Figure~\ref{Figure: example}. These sequences of events, connected in a temporal order, form linear event chains, which can be viewed as a form of linear representation of the progress of described events in a narrative.
%\muhao{Since this section is about ``event chains and narratives'', here it needs another sentence to say what's the relationship between the event chain and a narrative (Say, the event chain can be viewed as a representation of the progress of described events in a narrative.)}

However, %not all 
events %have the same importance in our understanding of the text. 
that co-occur or are described together in an article are often not equally important.
Some events are \textit{salient} -- they are relevant to the main topic %in that they provide key information of a story or connect other entities and events
of a text context, or play essential roles with regard to the central goal of an event chain~\citep{liu-etal-2018-automatic-event,chen-etal-2020-trying}. For example, from the paragraph in Figure~\ref{Figure: example}, the events \textit{detained}, \textit{alerted} and \textit{questioned} %are considered salient
are salient components that constitute the progress of a described story. Other events may describe how the salient events came to be known or involve other events that happened alongside salient events but are not critical to understanding the core actions of the story. In Figure~\ref{Figure: example}, \textit{said} and \textit{establish} are not salient. The goal of event salience detection systems~\citep{liu-etal-2018-automatic-event,jindal-etal-2020-killed} is to identify events %that are most relevant,
that are essential components of the narrative,
which can help us filter out trivial or distracting event mentions from event chains.

From another related %but different direction
perspective, modeling discourse structures of the article is also helpful for understanding which parts of the text directly report on main events and which parts serve other supportive roles. Discourse profiling~\citep{choubey-etal-2020-discourse} seeks to analyze the %function and role
functional role of each sentence, which is different from the %event-level classification of salience detection
event-level prediction of salience. \citet{van1988news}'s theory of news discourse outlines an ontology of sentence types that are seen in most news articles. According to the theory, the first sentence in Figure~\ref{Figure: example} is a \textit{Main Event} sentence, while the second is an \textit{Evaluation} sentence. We may classify the members of \citeauthor{van1988news}'s ontology into types that are core to understanding the event sequence of a story (e.g. \textit{Main Event}) and types that are not (e.g. \textit{Evaluation}), and further refine extracted event sequences. %It also provides an opportunity to incorporate event-level salience information in sentence-level discourse classification, to achieve a better filtering effect.

Though salience and discourse structures represent different perspectives of narrative analysis, they can be concurrently modeled in an event chain extraction system, leading to more effective filtering of mined event chain representations.

\begin{figure*}[t]
    \centering
    \includegraphics[width=\textwidth]{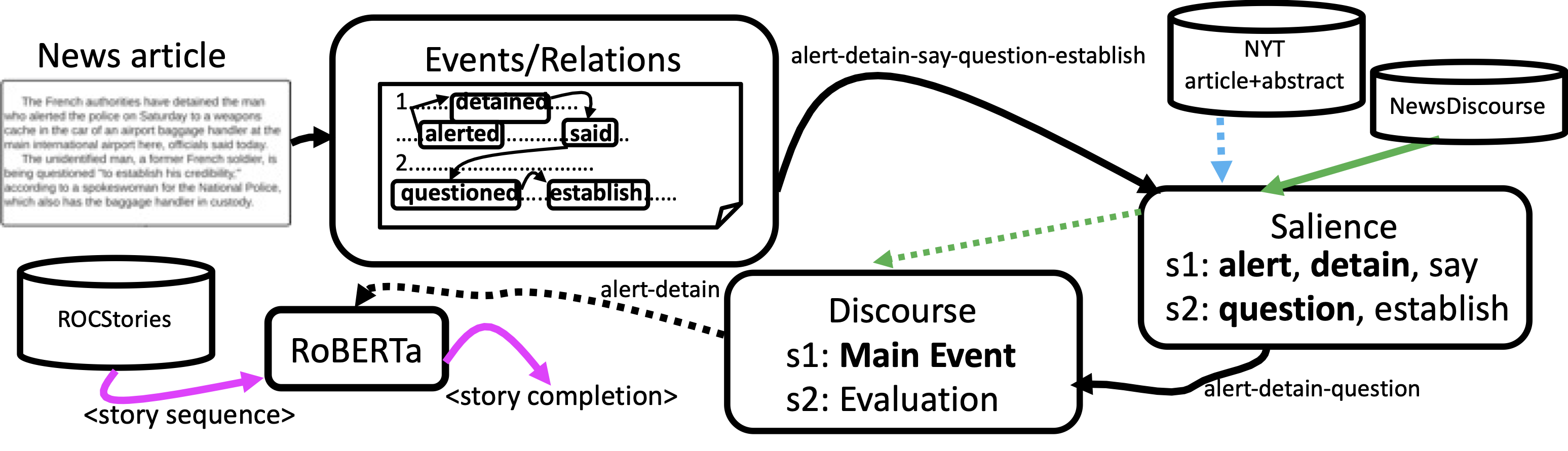}
    \caption{System diagram of our approach along with an example. Solid lines indicate inference and dotted lines indicate training. Colors separate different data streams. Events in temporal order are extracted from the news article from Figure~\ref{Figure: example}. Salience detection, trained on (article, abstract) pairs, filters out unimportant individual events from the example as well as labeled news discourse training data. Our salience-aware discourse parsing model removes sentences that do not contribute to the direct story line. The important event chain is used to fine-tune a masked language model, which is used to predict story completion. A similar pipeline is used for the TORQUE task. %The model is then fine-tuned for a particular task as normal.
    }
    \label{Figure: pipeline}  
\end{figure*}

\section{Method}

To obtain the %intended
interesting event chains from a document, we first use the TEAR tool by~\citet{han-etal-2019-joint} to generate a temporal relation graph. Then we apply different levels of filtering on the extracted events, which correspond to the nodes in the event graph structure. Finally, we extract linear chains of events from the filtered graph structure by following along the directed edges. In our evaluation, we will use these linear chains instead of raw text to pre-train language models such as RoBERTa~\citep{liu2019roberta}. Our overall pipeline is shown in Figure~\ref{Figure: pipeline}.
Details of each step in the pipeline are described as follows.

\subsection{Event Chain Extraction}
We adopt the joint structured event-relation extraction model %architecture 
from~\citet{han-etal-2019-joint}. It uses pre-trained BERT~\citep{devlin-etal-2019-bert}
%word 
embedding vectors of the input text,
which are further fed into an RNN-based scoring function for both event and relation extraction. During the end-to-end training, a MAP inference framework sets up a global objective function using local scoring functions to get the optimal assignment of event and relation labels. To ensure that we obtain globally coherent assignments, several logical constraints are specified including event-relation consistency and symmetry/transitivity consistency, so that the output event graph is %valid
logically consistent. This end-to-end %architecture
model extracts both events and event relations simultaneously.

We restrict the event relation labels to \textit{BEFORE} and \textit{NONE}. We can regard each \textit{BEFORE} relation as a directed edge in the temporal relation graph that points from the previous event to the next. 
Finally, to extract linear chains from this directed graph, we repeatedly choose the starting node in a topological order, and start a walk along the directed edge to obtain a maximum chain.

\subsection{Event Salience Filtering}
\label{salience}
For the event salience detection task, we follow the Kernel-based Centrality Estimation model by~\citet{liu-etal-2018-automatic-event}. This model combines the various event salience features such as frequency, sentence location and average cosine similarity with other events or entities, with Gaussian kernels that model event-event and entity-event interactions. For the final salience score we apply an additional sigmoid function and use binary cross-entropy loss for training. In this way, we formulate the task as event-level binary classification.

We train the event salience model on the New York Times (NYT) Annotated Corpus~\citep{linguistic2008new}, a collection of news articles with expert-written abstracts. During training, we use the frame-based event mention annotation by~\citet{liu-etal-2018-automatic-event}, and for inference on new articles, we extract the event mentions using TEAR. Each event mention is labeled as salient if its lemma appears in the associated abstract. The trained salience model can then assign a salience score between 0 and 1 to each event extracted from a document, and we use a threshold of 0.5 to perform the final filtering of the events.

\subsection{Discourse Salience Filtering}
\label{salience-discourse}

For the discourse parsing model, we follow the hierarchical architecture by~\citet{choubey-etal-2020-discourse} to assign each sentence with one of the eight fine-grained content labels\footnote{Those include \textit{Main Event} (M1), \textit{Consequence} (M2), \textit{Previous Event} (C1), \textit{Current Context} (C2), \textit{Historical Event} (D1), \textit{Anecdotal Event} (D2), \textit{Evaluation} (D3), and \textit{Expectation} (D4).} modified from \citeauthor{van1988news}'s theory of news discourse~(\citeyear{van1988news}). The sentences that are labeled as either \textit{Main Event, Consequence, Previous Event} or \textit{Current Context} are considered relevant to the main sequence, and kept in the sentence-level filtering process.

\begin{figure}[t]
    \centering
    \includegraphics[width=\columnwidth]{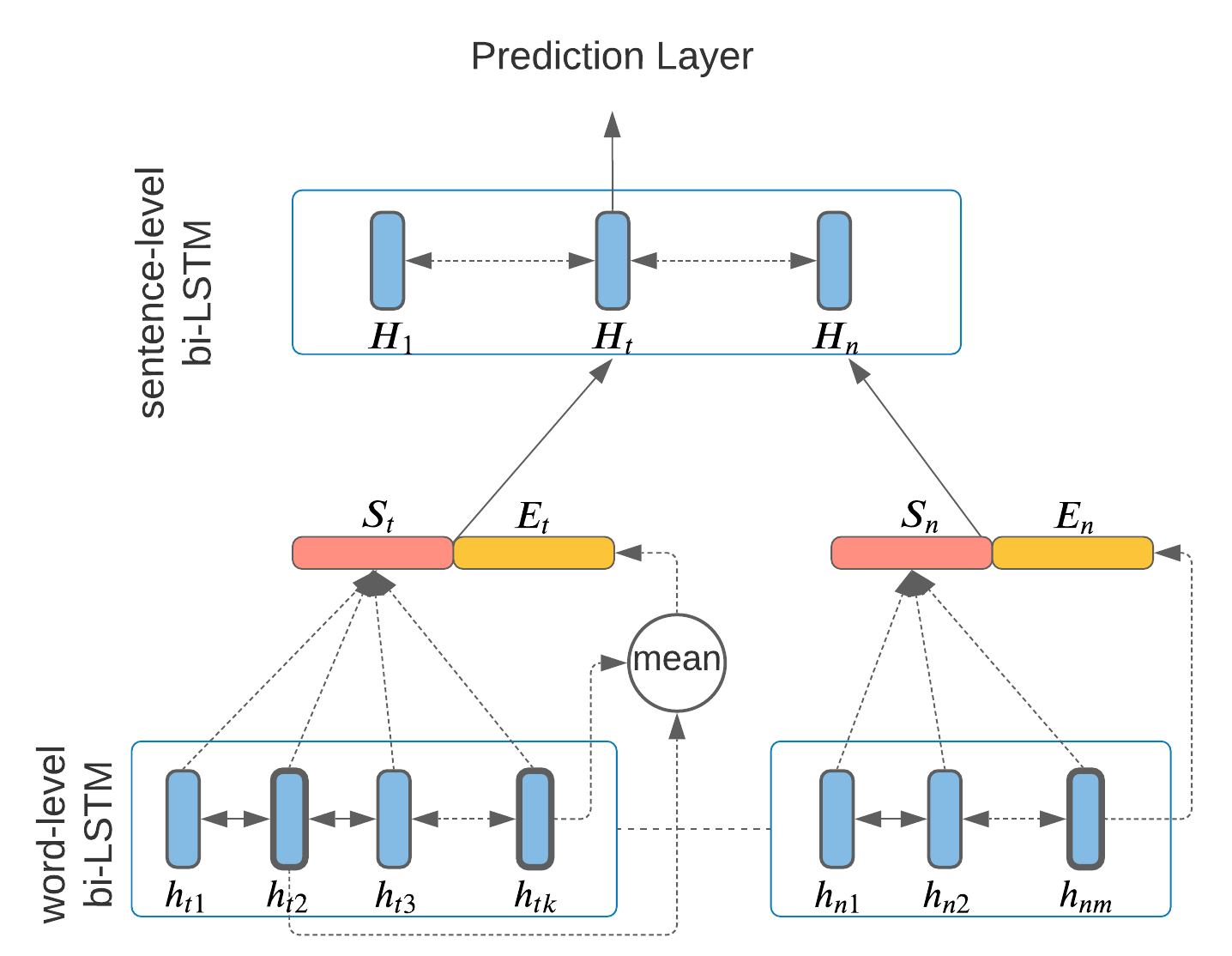}
    \caption{Architecture of salience-aware discourse parsing model. The addition of $E_t$ encoding for each sentence, which is an average of the salience event encodings, provides additional event-level salience information not seen in the original discourse parsing model.}
    \label{Figure: discourse}  
\end{figure}

One %naive
straightforward way to incorporate information from event salience in the filtering procedure is to first apply discourse filtering, and then keep only salient events from the discourse-filtered sentences, but this may propagate errors in the process and filter away too many events.
We instead %come up with using
 use event salience information to improve the performance on the discourse parsing task. As shown in Figure~\ref{Figure: discourse}, we modify the hierarchical bi-LSTM model from the work by~\citet{choubey-etal-2020-discourse}. The input is a document consisting of sentences $s_1, s_2, \dots, s_n$. Each sentence $s_t$, which is a sequence of tokens $(w_{t1},w_{t2},\dots,w_{tk})$, is first transformed to a sequence of ELMo~\citep{peters-etal-2018-deep} word representations, and then to hidden vectors $(h_{t1},h_{t2},\dots,h_{tk})$ in the word-level bi-LSTM layer. Then, we concatenate the original intermediate sentence encoding $S_t$ %which is
obtained from an soft-attention-weighted sum of the hidden vectors, with $E_t$ which is an average over the hidden vectors of salient events within the sentence. These concatenated vectors are fed into the sentence-level bi-LSTM layer to generate sentence encodings $H_1, \dots, H_t, \dots, H_n$, %and the final sentence representations and prediction layer follow from the original architecture.
on top of which the final prediction layer and softmax are stacked, following the original model architecture by~\citet{choubey-etal-2020-discourse}.
Specifically, we compute the document encoding $D$ as a soft-attention-weighted sum over the sentence encodings, and the final sentence representation of sentence $t$ is the concatenation of $H_t$ together with the element-wise product and difference between $H_t$ and $D$. The final sentence representation is used in a two-layer feed-forward neural network to make final prediction of the sentence's news discourse label.

We train the described model on the NewsDiscourse corpus with annotated content labels~\citep{choubey-etal-2020-discourse}, following their training setup. 
After we use our trained discourse parsing model on an input document, we filter the document down to only the salience-aware discourse-filtered sentences, which are sentences classified as one of \textit{Main Event, Consequence, Previous Event} and \textit{Current Context}%(corresponding to either the main content or context-informing content)
, and keep only events from these sentences. We also try keeping only salient events from the filtered sentences, but it leads to worse performance as shown in our experiments.

\subsection{Event Language Models}
Once we obtain the linear event chains after performing the extraction and filtering from a text dataset, we treat the sequence of event mentions in each chain as a sequential context that would be used for training or fine-tuning a language model. %For example, when training an RNN-based language model, the input layer takes a sequence of event mentions in order instead of a sequence of words in the raw text. 
For example, we can fine-tune a pre-trained Transformer language model based on the event chains (Section~\ref{sec:temporal}), or capture the sequences by training an RNN language model from scratch (Section~\ref{sec:narrative}).
%For event prediction and ordering prediction for the two tasks described in the next section.
Once we obtain such a language model, we seek to use it to help with narrative prediction by predicting the continuation of an event chain.
We can also leverage the fine-tuned model to support question answering regarding temporal orders of events.
%\muhao{I think another (perhaps short) subsection to describe how to obtain the event chain LM is also needed, in order to complete the method description. The two tasks depend on such an LM, whose details are missing.}
%\muhao{I'll try to modify this subsection a bit.}

\section{Experiments}

It is difficult to directly evaluate the quality of event chains in an intrinsic way. %as there is little consensus as to this definition. 
Some works on event schemas ask human annotators to rate qualities of generated chains~\citep{balasubramanian-etal-2013-generating,Weber_Balasubramanian_Chambers_2018,weber-etal-2018-hierarchical}, which can be subjective. We instead turn to extrinsic evaluation tasks that depend on implicitly understanding  typical sequences of meaningful events in order to be completed usefully.

TORQUE~\citep{ning-etal-2020-torque}, short for \textit{Temporal ORdering QUEstions}, is a machine reading comprehension %question answering 
benchmark that requests a model to answer questions regarding temporally specified events (e.g., ``What happened before the snow started?'') %given a text describing a sequence of events.
in a reference article. We hypothesize that models trained on unfiltered event chains are less likely to focus on the relevant events requested by the question, but this seeks to be improved by our filtered event chains.

ROCStories~\citep{mostafazadeh-etal-2016-corpus} is a narrative prediction dataset consisting of five-sentence short stories, where each sentence of a story contains a core event. A test set included with ROCStories contains %an alternate end to each story that is not relevant. 
two candidate endings to each partially complete story, where one of them is plausible.
While prior work has successfully leveraged event chains to infer the endings~\citep{chaturvedi-etal-2017-story},
we believe a model trained on relevant chains of events should be able to better distinguish the relevant ending from the irrelevant one than a model trained on event chains that contain irrelevant events. Other works such as~\citet{sun-etal-2019-improving} have directly tried to maximize performance on this corpus; we don't seek to directly compete with that work here, but rather use ROCStories as a means of establishing that our isolation of important events at the event language model level does positively influence event ending prediction, indicating that the sequences we find do indeed contain more relevant events.

In this section we verify the value of our event-filtering model on these tasks. We also evaluate our proposed discourse parsing model to show the usefulness of combining salience information at multiple levels.

\subsection{Answering Temporal Ordering Questions}\label{sec:temporal}

\stitle{Dataset}
The TORQUE dataset has 3.2k news snippets and 21.2k user-provided questions. We follow the original data split given by~\citet{ning-etal-2020-torque}, using the training set for fine-tuning the language model for reading comprehension, and the dev set for evaluation. Each question asks about specific temporal relationships of the events in a news passage, and requests a sequence of event %words
mentions from the passage that answer the question.

\begin{table*}[t]
\centering
\begin{tabular}{l|cc|cc|cc}
\hline
    \multirow{2}{*}{\textbf{Training Setting}} & \multicolumn{2}{c|}{\textbf{Textual order}} &
    \multicolumn{2}{c|}{\textbf{TEAR order}} & 
    \multicolumn{2}{c}{\textbf{TEAR (GPT-2)}} \\
    & F1 & EM & F1 & EM & F1 & EM\\ \hline\hline
   Baseline~\citep{ning-etal-2020-torque} & 75.5 & 50.1 & - & - & - & -\\\hline
   All events & 75.9 & 50.2 & 75.8 & 50.2 & 78.9 & 53.5\\
   Salient events & 76.2 & 50.7 & 76.0 & 50.6 & 79.5 & 54.5\\
   Discourse-filtered events & 76.2 & 50.6 & 76.3 & 50.7 & 79.4 & 54.3 \\
   Salience-aware discourse-filtered & 76.8 & 51.0 & 76.9 & 51.2 & \textbf{79.9} & \textbf{54.8}\\
   Salient + Salience-aware discourse-filtered & 76.4 & 50.6 & 76.5 & 50.9 & 79.5 & 54.6\\\hline 
\end{tabular}
\caption{Macro F1 and Exact-Match metrics (\%) on TORQUE dev set under different training settings. The baseline setting corresponds to the original best performing setting of RoBERTa-large used in~\citet{ning-etal-2020-torque} which does not use event chains, and the other settings refer to the different event-filtering methods. \textbf{Textual order} or \textbf{TEAR order} denotes if we construct the event chains in the order they appear in text or in the temporal order predicted by TEAR. The rightmost \textbf{TEAR (GPT-2)} column denotes the setting where we substitute RoBERTa-large with GPT-2.}
\label{table:torque table}
\end{table*}

\begin{table*}[t]
\centering
\begin{tabular}{l|ccc}
\hline
   \textbf{Training Setting}
    & Before & After & While \\ \hline\hline
   Baseline~\citep{ning-etal-2020-torque} & 81.2 & 82.3 & 76.9\\\hline
   All events & 82.9 & 83.7 & 77.5 \\
   Salient events & 83.3 & 84.2 & 77.8 \\
   Discourse-filtered events & 83.7 & 84.8 & 78.0  \\
   Salience-aware discourse-filtered & \textbf{84.4} & \textbf{85.3} & \textbf{78.4} \\ \hline 
\end{tabular}
\caption{Macro F1 scores (\%) on TORQUE dev set under different training settings, with a more detailed breakdown into ``standard questions''. Event chains are in the temporal order predicted by TEAR, and other training settings are the same as in Table~\ref{table:torque table}.}
\label{table:torque table detail}
\end{table*}

\stitle{Evaluation Details}
We follow
\citet{ning-etal-2020-torque} to fine-tune a RoBERTa-large~\citep{liu2019roberta} model on the training set of TORQUE, which has a perceptron layer that classifies whether each token in the passage is in the answer or not. The input to the model is the question followed by the passage, separated by the separator token. We follow the same approach, but further fine-tune the model on inputs containing the extracted (and possibly filtered) \textit{event chain}, rather than the entire passage.
%of the form ``\texttt{<s>} question \texttt{<sep>} event chain \texttt{</s>}''. 
For training, we use the same batch size and learning rate as the baseline approach by~\citet{ning-etal-2020-torque} in each experiment. We evaluate on the two standard metrics used in question answering, namely macro F1 and exact-match (EM), on the development set, comparing between models fine-tuned on filtered event chains and unfiltered chains, and, in Table~\ref{table:torque table}, report the average performance over the 3 training runs started from different random seeds.
As a comparison, we also fine-tune a GPT-2-based model~\citep{radford2019language}, which leads to better results as shown in the rightmost column in Table~\ref{table:torque table}.
%\muhao{What's the Baseline in Tables 1 and 2? Need to explicitly say the baseline is the best performing setting of RoBERTa-large in Ning-etal.}

\stitle{Results and Analysis}
As we see in Table~\ref{table:torque table}, we improve the performance on predictions of events over the baseline by fine-tuning on event chains constructed in various ways. Also, using filtered event chains gives better results than using unfiltered chains, regardless of the method of filtering. The model achieves the highest score when we use our salience-aware discourse filtering method on event chains, which shows the effectiveness of our approach of combining salience and discourse information. The last two rows of Table~\ref{table:torque table} show that keeping only salient events from the discourse-filtered sentences leads to worse results than keeping all events from the discourse-filtered sentences. We see the same trend in the last column, which suggests that the improvements from salience and discourse hold on GPT-2 as well as on RoBERTa fine-tuning. Comparing between the numbers from the two columns of ``textual order''\footnote{TEAR is used to extract events but the temporal order relations are ignored.} and ``TEAR order'' in the table, we see the order of event chains does not seem to affect the performance much, probably because the number of events per document is not very large.

\begin{figure*}[t]
    \centering
    \includegraphics[width=0.81\textwidth]{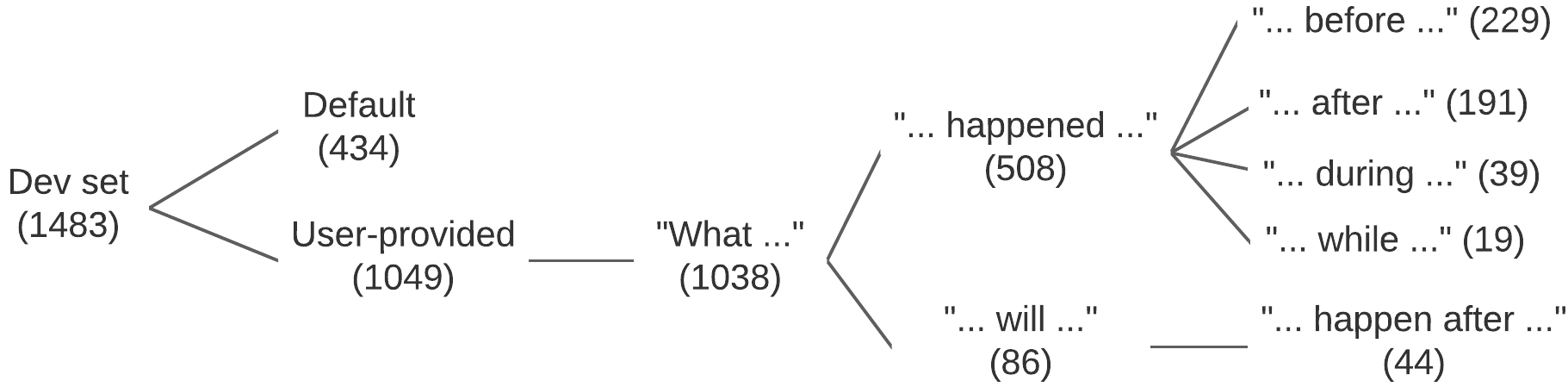}
    \caption{Distribution of questions in the TORQUE dev set, showing only the most frequent prefixes related to Before, After or Co-occuring relations. For example, a total of 229 questions start with ``What happened before'', which query the Before relation. 191+44 questions query the After relation, and 39+19 questions query the Co-occuring relation. We define these three categories of questions as ``standard'' type. This prefix matching is only a rough categorization of the question type.}
    \label{Figure:torque stats}  
\end{figure*}

To further illustrate the effect of the various filtering methods on temporal understanding, we perform a %rough 
breakdown of the questions in the dev set into different types, using prefix matching as shown in Figure~\ref{Figure:torque stats}. We define ``standard'' questions as those directly asking about Before, After or Co-occuring relations, without the additional challenges from the fuzziness of time scopes or non-factual modality, as mentioned by~\citet{ning-etal-2020-torque}. The default questions ask about which events had already happened, were ongoing, or were still in the future, which we regard as not in the ``standard'' category. After examining the F1 scores on these ``standard'' categories in Table~\ref{table:torque table detail}, we see that the model achieves a larger improvement with our filtering methods compared to on all questions, even though these are categories that already have a higher baseline performance.

\subsection{Narrative Prediction}\label{sec:narrative}

\stitle{Dataset}
We evaluate the effectiveness of our extracted event chains on narrative prediction using the ROCStories dataset~\citep{mostafazadeh-etal-2016-corpus}. 
The training set contains 98,161 five-sentence stories, and the development and test sets each consist of 1,871 instances of four-sentence stories together with a correct and an incorrect ending. The goal is to predict the correct ending sentence given the two candidate choices. %also known as the Story Cloze Test. 
We use the development set but not the training set for supervised evaluation, following previous approaches~\citep{chaturvedi-etal-2017-story,srinivasan-etal-2018-simple}, as detailed below. 

\stitle{Evaluation Details}
Since %each story in the ROCStories dataset consists of five relatively simple sentences each containing a core event
each of the five sentences in a ROCStories article contains a core event, we convert the task to the prediction of the next event given the current sequence of events.
%For event chain extraction,
We use the NYT corpus following \citet{chaturvedi-etal-2017-story}, %which is an external dataset, 
and run our event extraction and filtering methods to obtain event chains from the news articles. We then break the obtained event chains into sequences of five events' length, and train a simple bi-LSTM masked language model on these sequences.
This seeks to compare with the best performing method using the same type of features \cite{chaturvedi-etal-2017-story}.
For evaluation on the Story Cloze Test, we choose the ending that contains the event with higher probability of occurring next given the four previous ones, according to the model. As a comparison, we also use the event chains to fine-tune a RoBERTa-large model. %for masked language modeling. 
In this way we are performing an \textit{unsupervised} evaluation in the sense that we are not learning directly from labeled sentences in ROCStories, and we are comparing the performance of %the trained model based on whether and how the input event chains are filtered
models trained/fine-tuned on event chains obtained from different filtering methods. We also present results using the standard supervised setting~\citep{chaturvedi-etal-2017-story,srinivasan-etal-2018-simple}, i.e., %rather than using the event language model's own probability of which of two events is more likely, we directly train a binary classifier on top,
on top of the language model, we train a binary classifier using the correct and incorrect endings from the development set as positive and negative examples respectively.

\stitle{Results and Analysis}
\begin{table*}[t]
\centering
\setlength{\tabcolsep}{3pt}
\begin{tabular}{l|cc|cc}
\hline
    \multirow{2}{*}{\textbf{Training Setting}} & \multicolumn{2}{c|}{\textbf{Textual order}} &
    \multicolumn{2}{c}{\textbf{TEAR order}} \\
    & unsupervised & supervised & unsupervised & supervised \\ \hline\hline
    Event-sequence SemLM~\citep{chaturvedi-etal-2017-story} & - & 71.6 & - & -\\
%    val-LS-skip~\citep{srinivasan-etal-2018-simple} & - & 76.5 & - & -\\
    \hline
   All events & 61.3 & 72.5 & 63.2 & 73.3\\
   Salient events & 64.1 & 73.4 & 63.6 & 73.8 \\
   Discourse-filtered events & 64.8 & 73.6 & 63.8 & 74.1 \\
   Salience-aware discourse-filtered & 66.2 & 74.2 & 65.8 & 74.5\\
   Salient + Salience-aware discourse-filtered & 65.0 & 73.9 & 64.4 & 74.2\\\hline
   Salience-aware discourse-filtered (RoBERTa) & \textbf{70.4} & \textbf{76.7} & \textbf{70.6} & \textbf{76.4}\\\hline
\end{tabular}
\caption{Accuracy (\%) on ROCStories test set under different training settings. \textbf{Textual order} or \textbf{TEAR order} denotes if we construct the event chains from the NYT corpus in the order they appear in text or in the temporal order predicted by TEAR. The ``unsupervised'' mode only trains on sequences from the NYT documents, where the ``supervised'' mode also uses the development set of ROCStories. 
The baseline setting is a comparable logistic regression model trained on event-sequence modeling features only~\citep{chaturvedi-etal-2017-story}.}
%, and the model that classifies the ending given the skip-thought embedding of the last sentence by~\citet{srinivasan-etal-2018-simple}.

%\muhao{Modifications: use \cite{chaturvedi-etal-2017-story}'s event chain LM as a baseline. Add Transformer-based results. 
%and if those outperforms \citet{srinivasan-etal-2018-simple}, and it as a baseline well.}
%\muhao{Not good to present in this way if we cannot outperform \cite{srinivasan-etal-2018-simple}.  \cite{srinivasan-etal-2018-simple} is not using the same kind of features as this work, right? Perhaps what we should put there is the best performing work that uses narrative event chains.} \jon{I think we can re-emphasize that we are simply trying to evaluate the value of filtering events, which is clearly established here.}%\jon{To be clear, when using TEAR order here, is this just for training data or is TEAR ordering used for evaluation too? I would think that for evaluation/supervised learning we could just use the events as they appear since ROCStories is not very convoluted w/r/t sentence ordering.}
\label{table:lstm table}
\end{table*}
From Table~\ref{table:lstm table}, we can see the improvement of prediction accuracy when we feed filtered event chains to our model compared with using unfiltered chains. Our salience-aware discourse-filtered event chains, compared among all event chain-based training settings, provide the best results in both the unsupervised and supervised settings, regardless of the order of extracting the raw event chains. %In the supervised setting, our LSTM-based approaches do not achieve better result than the Hidden Coherence Model used by~\citet{chaturvedi-etal-2017-story} which also utilized language modeling on event chains but other semantic features.
Under the unsupervised setting in which the model has not leveraged any supervision from the development set, and with event chains extracted in textual order, we see that event salience filtering and discourse filtering provide a 2.8\% and 3.5\% improvement on their own, respectively, and the largest improvement of 4.9\% occurs when we use our proposed salience-aware discourse filtering. We see a similar trend under other evaluation settings, that combining event-level and discourse-level salience filtering leads to better performance, though the improvements are not as significant. %\muhao{This part of discussion needs to be expanded. Just say ``nearly 5\%'' is too vague. A more detailed discussion should consider how much each part of the incorporated filtering strategy contributes.}
This confirms that even without additional supervision from labeled data, we can utilize salience-aware and discourse-aware filtering to improve the relevance of event chains for better narrative prediction results. From the last row of Table~\ref{table:lstm table}, we see that our RoBERTa-based masked language model trained on the best produced event chains is able to further improve the performance in the supervised evaluation. 

\subsection{Salience-Aware Discourse Parsing}

\begin{table*}[t]
\centering
\begin{tabular}{l|cccccccc|c|c}
\hline
   \textbf{Models} & M1 & M2 & C1 & C2 & D1 & D2 & D3 & D4 & \textbf{Macro} & \textbf{Micro}\\ \hline
Baseline & 49.6 & 27.9 & 22.5 & 58.1 & 64.1 & 48.1 & 67.4 & 57.6 & 54.4 & 60.9\\
Ours     & 55.1 & 30.7 & 26.3 & 64.4 & 63.8 & 48.5 & 67.8 & 58.0 & 58.2 & 62.8\\ \hline 
\end{tabular}
\caption{Per-class and average F1 scores (\%) of proposed salience-aware discourse parsing model, in the same evaluation setting as~\citet{choubey-etal-2020-discourse} (average of 10 runs with random seeds). Baseline model is the best performing one (\textbf{Document LSTM + Document Encoding}).}
\label{table:discourse}
\end{table*}

As %an intrinsic evaluation
a case study of our proposed discourse parsing model from Section~\ref{salience-discourse}, we also compare our performance on the discourse type classification task with the baseline model from~\citet{choubey-etal-2020-discourse}, as shown in Table~\ref{table:discourse}. We see that our model surpasses the baseline model in both macro and micro F1 scores. Looking at the F1 score of each specific discourse type, we achieve the greatest improvement in the classification of type M1 (\textit{Main Event}) and C2 (\textit{Current Context}), with an increase of 5.5\% and 6.3\% respectively. This suggests that introducing salience awareness in the discourse parsing model indeed leads to better prediction accuracy, especially in categories that we expect to be most relevant to the topic of a document.

\section{Related Work}
%This work follows in the tradition of \citet{chambers-jurafsky-2008-unsupervised}, who identify the value of event schemas, or sequences of important events that, when taken together, comprise larger complex events, at predicting
We discuss three relevant research topics.

\stitle{Event Chains}
%Numerous recent works have identified the importance of event chains.
Much research effort has been made to extract and process event chains.
\citet{chambers-jurafsky-2008-unsupervised} pioneered the modern interest and %extracted frequently occurring sequences of via unsupervised approach.
modeled the co-occurrence of events in narrative chains based on their PMI.
\citet{10.1145/2187836.2187958} and \citet{10.1145/2433396.2433431} %investigate automated extraction and generalization of event chains
extended such unsupervised event chain modeling to cross-document scenarios
and used the technology for news prediction and timeline construction. \citet{berant-etal-2014-modeling} 
%model rich structures that represent complex relations
extracted relations among events and entities in biological processes to help solve a biological reading comprehension task using a structure matching method. More recently, \citet{chen-etal-2020-trying} attempted to infer the type of action and object associated with an event chain, which required recognition of the goal or intention extracted from the chain. 
\citet{zhang-etal-2020-analogous} used a probabilistic graphical model to capture common patterns from analogous event chains, and induced new chains from those patterns. These works do not explore salience or discourse structures for event chains.

\newpage

%While the aforementioned predictive models do not explore salience or discourse structures for event chains, such factors are essential to predictive tasks on narratives and temporal understanding, which is what we focus on in this work.

\stitle{Discourse and Salience}
There have been recent interests in event salience identification. \citet{choubey-etal-2018-identifying} studied how to find the dominant event(s) in a news article using mined event coreference relations. \citet{liu-etal-2018-automatic-event} proposed a sequence tagging model for event salience detection, %which was applied on more general domains, 
and also contributed a large-scale event salience corpus based on NYT. \citet{jindal-etal-2020-killed} tried to further capture representation of events and their interactions, and evaluated the modeling of event salience on extractive summarization.

Recent studies on discourse structures, especially for news articles, were built on Van Dijk's theory~\citep{van1988news}. For example, \citet{yarlott-etal-2018-identifying} annotated a dataset of discourse structures, which viewed paragraphs as units of annotations, and developed models to predict discourse labels. \citet{choubey-etal-2020-discourse} instead created a sentence-level discourse structure corpus spanning different domains and sources, which was more helpful for a fine-grained identification of sentences relevant to the main topic. Insights in those studies constitute the two aspects of salience awareness that are characterized in our method, i.e., event-level and discourse-level salience.

\stitle{Event-Centric Language Models}
%Since masked language models became popular
Another line of research focuses on language modeling for capturing sequences of events.
Many works in this line extract raw event chains and directly train a neural language model for narrative prediction~\citep{chaturvedi-etal-2017-story,peng-etal-2019-knowsemlm} or script generation~\citep{rudinger-etal-2015-script,10.5555/3016100.3016293,weber-etal-2018-hierarchical}.
%~\citep{devlin-etal-2019-bert,liu2019roberta}, there have been works exploring how to represent non-pure text like events and use them in the fine-tuning procedure.
Besides directly training,
\citet{10.1145/3397271.3401173} proposed a unified fine-tuning architecture, where a masked language model was explicitly fine-tuned on the representations of event chains to model event elements. 
\citet{li-etal-2020-connecting} proposed to learn to induce event schemas using an auto-regressive language model trained on salient paths on an event-event relation graph, which was also an attempt to reduce noise in constructing event schema.
While that work uses a different kind of data structure, our works are in agreement  on the importance of incorporating salience when fine-tuning an event-centric language model. 
Specifically, in comparison to prior works that used the language model for narrative prediction tasks \citep{chaturvedi-etal-2017-story,peng-etal-2019-knowsemlm}, we show that salience-awareness is an important factor to tackle such tasks.

%\citet{wadden-etal-2019-entity} build span representations on top of multi-sentence BERT encodings, where text spans are constructed by concatenating the tokens representing their left and right endpoints, together with a learned span width embedding.

\section{Conclusion}
We propose an event chain extraction %and filtering 
pipeline, which %is discourse-aware and salience-aware
leverages both salience identification and discourse profiling to filter out distracting events. We demonstrate the effectiveness of the approach by using the produced event chains to train/fine-tune language models, which leads to improved performance on temporal understanding of events and narrative prediction. Our case study on salience-aware discourse parsing shows the advantage of combining event-level and sentence-level salience information. We plan to use these event chain patterns on other narrative understanding and generation tasks, such as constrained story generation \cite{peng-etal-2018-towards}, event script generation \cite{zhang-etal-2020-analogous,lyu-etal-2021-goal}, and implicit event prediction \cite{lin-etal-2021-conditional,zhou-etal-2021-temporal}.

\section*{Ethical Considerations}
This work does not present any direct societal consequence. The proposed method aims at providing high-quality extraction of event chains from documents with awareness of salience and discourse structures, but is only evaluated on English data and uses western notions of both salience and news discourse; event sequence extraction using data from other languages and cultures may not benefit from the methods shown here. The extracted event chain representations benefit narrative understanding and temporal understanding of events. 
Yet, real-world open source articles may include societal biases. 
%Reasoning over biased UKGs may support or magnify those biases. 
Extracting event chains from articles with such biases may potentially propagate the bias into acquired knowledge representation.
While not specifically addressed in this work, the ability to incorporate salience and discourse-awareness could be one way to mitigate bias.

\section*{Acknowledgements}
We appreciate the anonymous reviewers for their insightful comments and suggestions.

This research is based upon work supported by DARPA's KAIROS program, Contract FA8750-19-2-0500. The views and conclusions contained herein are those of the authors and should not be interpreted as necessarily representing the official policies or endorsements, either expressed or implied, of DARPA or the U.S. Government. The U.S. Government is authorized to reproduce and distribute reprints for Governmental purposes notwithstanding any copyright annotation thereon.

\bibliography{anthology,custom}

\begin{thebibliography}{42}
\expandafter\ifx\csname natexlab\endcsname\relax\def\natexlab#1{#1}\fi

\bibitem[{Balasubramanian et~al.(2013)Balasubramanian, Soderland, {Mausam}, and
  Etzioni}]{balasubramanian-etal-2013-generating}
Niranjan Balasubramanian, Stephen Soderland, {Mausam}, and Oren Etzioni. 2013.
\newblock \href {https://aclanthology.org/D13-1178} {Generating coherent event
  schemas at scale}.
\newblock In \emph{Proceedings of the 2013 Conference on Empirical Methods in
  Natural Language Processing}, pages 1721--1731, Seattle, Washington, USA.
  Association for Computational Linguistics.

\bibitem[{Berant et~al.(2014)Berant, Srikumar, Chen, Vander~Linden, Harding,
  Huang, Clark, and Manning}]{berant-etal-2014-modeling}
Jonathan Berant, Vivek Srikumar, Pei-Chun Chen, Abby Vander~Linden, Brittany
  Harding, Brad Huang, Peter Clark, and Christopher~D. Manning. 2014.
\newblock \href {https://doi.org/10.3115/v1/D14-1159} {Modeling biological
  processes for reading comprehension}.
\newblock In \emph{Proceedings of the 2014 Conference on Empirical Methods in
  Natural Language Processing ({EMNLP})}, pages 1499--1510, Doha, Qatar.
  Association for Computational Linguistics.

\bibitem[{Chambers and Jurafsky(2008)}]{chambers-jurafsky-2008-unsupervised}
Nathanael Chambers and Dan Jurafsky. 2008.
\newblock \href {https://www.aclweb.org/anthology/P08-1090} {Unsupervised
  learning of narrative event chains}.
\newblock In \emph{Proceedings of ACL-08: HLT}, pages 789--797, Columbus, Ohio.
  Association for Computational Linguistics.

\bibitem[{Chaturvedi et~al.(2017)Chaturvedi, Peng, and
  Roth}]{chaturvedi-etal-2017-story}
Snigdha Chaturvedi, Haoruo Peng, and Dan Roth. 2017.
\newblock \href {https://doi.org/10.18653/v1/D17-1168} {Story comprehension for
  predicting what happens next}.
\newblock In \emph{Proceedings of the 2017 Conference on Empirical Methods in
  Natural Language Processing}, pages 1603--1614, Copenhagen, Denmark.
  Association for Computational Linguistics.

\bibitem[{Chen et~al.(2020)Chen, Zhang, Wang, and Roth}]{chen-etal-2020-trying}
Muhao Chen, Hongming Zhang, Haoyu Wang, and Dan Roth. 2020.
\newblock \href {https://doi.org/10.18653/v1/2020.conll-1.43} {What are you
  trying to do? semantic typing of event processes}.
\newblock In \emph{Proceedings of the 24th Conference on Computational Natural
  Language Learning}, pages 531--542, Online. Association for Computational
  Linguistics.

\bibitem[{Cheng and Erk(2018)}]{cheng2018implicit}
Pengxiang Cheng and Katrin Erk. 2018.
\newblock \href {https://www.aclweb.org/anthology/N18-1076} {Implicit argument
  prediction with event knowledge}.
\newblock In \emph{Proceedings of the 2018 Conference of the North American
  Chapter of the Association for Computational Linguistics: Human Language
  Technologies, Volume 1 (Long Papers)}, pages 831--840.

\bibitem[{Choubey et~al.(2020)Choubey, Lee, Huang, and
  Wang}]{choubey-etal-2020-discourse}
Prafulla~Kumar Choubey, Aaron Lee, Ruihong Huang, and Lu~Wang. 2020.
\newblock \href {https://doi.org/10.18653/v1/2020.acl-main.478} {Discourse as a
  function of event: Profiling discourse structure in news articles around the
  main event}.
\newblock In \emph{Proceedings of the 58th Annual Meeting of the Association
  for Computational Linguistics}, pages 5374--5386, Online. Association for
  Computational Linguistics.

\bibitem[{Choubey et~al.(2018)Choubey, Raju, and
  Huang}]{choubey-etal-2018-identifying}
Prafulla~Kumar Choubey, Kaushik Raju, and Ruihong Huang. 2018.
\newblock \href {https://doi.org/10.18653/v1/N18-2055} {Identifying the most
  dominant event in a news article by mining event coreference relations}.
\newblock In \emph{Proceedings of the 2018 Conference of the North {A}merican
  Chapter of the Association for Computational Linguistics: Human Language
  Technologies, Volume 2 (Short Papers)}, pages 340--345, New Orleans,
  Louisiana. Association for Computational Linguistics.

\bibitem[{Devlin et~al.(2019)Devlin, Chang, Lee, and
  Toutanova}]{devlin-etal-2019-bert}
Jacob Devlin, Ming-Wei Chang, Kenton Lee, and Kristina Toutanova. 2019.
\newblock \href {https://doi.org/10.18653/v1/N19-1423} {{BERT}: Pre-training of
  deep bidirectional transformers for language understanding}.
\newblock In \emph{Proceedings of the 2019 Conference of the North {A}merican
  Chapter of the Association for Computational Linguistics: Human Language
  Technologies, Volume 1 (Long and Short Papers)}, pages 4171--4186,
  Minneapolis, Minnesota. Association for Computational Linguistics.

\bibitem[{Han et~al.(2019)Han, Ning, and Peng}]{han-etal-2019-joint}
Rujun Han, Qiang Ning, and Nanyun Peng. 2019.
\newblock \href {https://doi.org/10.18653/v1/D19-1041} {Joint event and
  temporal relation extraction with shared representations and structured
  prediction}.
\newblock In \emph{Proceedings of the 2019 Conference on Empirical Methods in
  Natural Language Processing and the 9th International Joint Conference on
  Natural Language Processing (EMNLP-IJCNLP)}, pages 434--444, Hong Kong,
  China. Association for Computational Linguistics.

\bibitem[{Jindal et~al.(2020)Jindal, Deutsch, and
  Roth}]{jindal-etal-2020-killed}
Disha Jindal, Daniel Deutsch, and Dan Roth. 2020.
\newblock \href {https://doi.org/10.18653/v1/2020.coling-main.10} {Is killed
  more significant than fled? a contextual model for salient event detection}.
\newblock In \emph{Proceedings of the 28th International Conference on
  Computational Linguistics}, pages 114--124, Barcelona, Spain (Online).
  International Committee on Computational Linguistics.

\bibitem[{Li et~al.(2020)Li, Zeng, Lin, Cho, Ji, May, Chambers, and
  Voss}]{li-etal-2020-connecting}
Manling Li, Qi~Zeng, Ying Lin, Kyunghyun Cho, Heng Ji, Jonathan May, Nathanael
  Chambers, and Clare Voss. 2020.
\newblock \href {https://doi.org/10.18653/v1/2020.emnlp-main.50} {Connecting
  the dots: Event graph schema induction with path language modeling}.
\newblock In \emph{Proceedings of the 2020 Conference on Empirical Methods in
  Natural Language Processing (EMNLP)}, pages 684--695, Online. Association for
  Computational Linguistics.

\bibitem[{Lin et~al.(2021)Lin, Chambers, and
  Durrett}]{lin-etal-2021-conditional}
Shih-Ting Lin, Nathanael Chambers, and Greg Durrett. 2021.
\newblock \href {https://doi.org/10.18653/v1/2021.acl-long.555} {Conditional
  generation of temporally-ordered event sequences}.
\newblock In \emph{Proceedings of the 59th Annual Meeting of the Association
  for Computational Linguistics and the 11th International Joint Conference on
  Natural Language Processing (Volume 1: Long Papers)}, pages 7142--7157,
  Online. Association for Computational Linguistics.

\bibitem[{Lin et~al.(2020)Lin, Ji, Huang, and Wu}]{lin-etal-2020-joint}
Ying Lin, Heng Ji, Fei Huang, and Lingfei Wu. 2020.
\newblock \href {https://doi.org/10.18653/v1/2020.acl-main.713} {A joint neural
  model for information extraction with global features}.
\newblock In \emph{Proceedings of the 58th Annual Meeting of the Association
  for Computational Linguistics}, pages 7999--8009, Online. Association for
  Computational Linguistics.

\bibitem[{Liu et~al.(2019)Liu, Ott, Goyal, Du, Joshi, Chen, Levy, Lewis,
  Zettlemoyer, and Stoyanov}]{liu2019roberta}
Yinhan Liu, Myle Ott, Naman Goyal, Jingfei Du, Mandar Joshi, Danqi Chen, Omer
  Levy, Mike Lewis, Luke Zettlemoyer, and Veselin Stoyanov. 2019.
\newblock {RoBERTa}: A robustly optimized {BERT} pretraining approach.
\newblock \emph{arXiv preprint arXiv:1907.11692}.

\bibitem[{Liu et~al.(2018)Liu, Xiong, Mitamura, and
  Hovy}]{liu-etal-2018-automatic-event}
Zhengzhong Liu, Chenyan Xiong, Teruko Mitamura, and Eduard Hovy. 2018.
\newblock \href {https://doi.org/10.18653/v1/D18-1154} {Automatic event
  salience identification}.
\newblock In \emph{Proceedings of the 2018 Conference on Empirical Methods in
  Natural Language Processing}, pages 1226--1236, Brussels, Belgium.
  Association for Computational Linguistics.

\bibitem[{Lyu et~al.(2021)Lyu, Zhang, and Callison-Burch}]{lyu-etal-2021-goal}
Qing Lyu, Li~Zhang, and Chris Callison-Burch. 2021.
\newblock \href {https://arxiv.org/pdf/2107.13189} {Goal-oriented script
  construction}.
\newblock In \emph{Proceedings of the 14th International Conference on Natural
  Language Generation}.

\bibitem[{Mostafazadeh et~al.(2016)Mostafazadeh, Chambers, He, Parikh, Batra,
  Vanderwende, Kohli, and Allen}]{mostafazadeh-etal-2016-corpus}
Nasrin Mostafazadeh, Nathanael Chambers, Xiaodong He, Devi Parikh, Dhruv Batra,
  Lucy Vanderwende, Pushmeet Kohli, and James Allen. 2016.
\newblock \href {https://doi.org/10.18653/v1/N16-1098} {A corpus and cloze
  evaluation for deeper understanding of commonsense stories}.
\newblock In \emph{Proceedings of the 2016 Conference of the North {A}merican
  Chapter of the Association for Computational Linguistics: Human Language
  Technologies}, pages 839--849, San Diego, California. Association for
  Computational Linguistics.

\bibitem[{Ning et~al.(2020)Ning, Wu, Han, Peng, Gardner, and
  Roth}]{ning-etal-2020-torque}
Qiang Ning, Hao Wu, Rujun Han, Nanyun Peng, Matt Gardner, and Dan Roth. 2020.
\newblock \href {https://doi.org/10.18653/v1/2020.emnlp-main.88} {{TORQUE}: A
  reading comprehension dataset of temporal ordering questions}.
\newblock In \emph{Proceedings of the 2020 Conference on Empirical Methods in
  Natural Language Processing (EMNLP)}, pages 1158--1172, Online. Association
  for Computational Linguistics.

\bibitem[{Ostermann et~al.(2019)Ostermann, Roth, and
  Pinkal}]{ostermann-etal-2019-mcscript2}
Simon Ostermann, Michael Roth, and Manfred Pinkal. 2019.
\newblock \href {https://doi.org/10.18653/v1/S19-1012} {{MCS}cript2.0: A
  machine comprehension corpus focused on script events and participants}.
\newblock In \emph{Proceedings of the Eighth Joint Conference on Lexical and
  Computational Semantics (*{SEM} 2019)}, pages 103--117, Minneapolis,
  Minnesota. Association for Computational Linguistics.

\bibitem[{Peng et~al.(2019)Peng, Ning, and Roth}]{peng-etal-2019-knowsemlm}
Haoruo Peng, Qiang Ning, and Dan Roth. 2019.
\newblock \href {https://doi.org/10.18653/v1/K19-1051} {{K}now{S}em{LM}: A
  knowledge infused semantic language model}.
\newblock In \emph{Proceedings of the 23rd Conference on Computational Natural
  Language Learning (CoNLL)}, pages 550--562, Hong Kong, China. Association for
  Computational Linguistics.

\bibitem[{Peng et~al.(2018)Peng, Ghazvininejad, May, and
  Knight}]{peng-etal-2018-towards}
Nanyun Peng, Marjan Ghazvininejad, Jonathan May, and Kevin Knight. 2018.
\newblock \href {https://doi.org/10.18653/v1/W18-1505} {Towards controllable
  story generation}.
\newblock In \emph{Proceedings of the First Workshop on Storytelling}, pages
  43--49, New Orleans, Louisiana. Association for Computational Linguistics.

\bibitem[{Peters et~al.(2018)Peters, Neumann, Iyyer, Gardner, Clark, Lee, and
  Zettlemoyer}]{peters-etal-2018-deep}
Matthew Peters, Mark Neumann, Mohit Iyyer, Matt Gardner, Christopher Clark,
  Kenton Lee, and Luke Zettlemoyer. 2018.
\newblock \href {https://doi.org/10.18653/v1/N18-1202} {Deep contextualized
  word representations}.
\newblock In \emph{Proceedings of the 2018 Conference of the North {A}merican
  Chapter of the Association for Computational Linguistics: Human Language
  Technologies, Volume 1 (Long Papers)}, pages 2227--2237, New Orleans,
  Louisiana. Association for Computational Linguistics.

\bibitem[{Pichotta and Mooney(2016)}]{10.5555/3016100.3016293}
Karl Pichotta and Raymond~J. Mooney. 2016.
\newblock Learning statistical scripts with {LSTM} recurrent neural networks.
\newblock In \emph{Proceedings of the Thirtieth AAAI Conference on Artificial
  Intelligence}, AAAI'16, page 2800–2806. AAAI Press.

\bibitem[{Radford et~al.(2019)Radford, Wu, Child, Luan, Amodei, and
  Sutskever}]{radford2019language}
Alec Radford, Jeff Wu, Rewon Child, David Luan, Dario Amodei, and Ilya
  Sutskever. 2019.
\newblock \href
  {https://d4mucfpksywv.cloudfront.net/better-language-models/language-models.pdf}
  {Language models are unsupervised multitask learners}.

\bibitem[{Radinsky et~al.(2012)Radinsky, Davidovich, and
  Markovitch}]{10.1145/2187836.2187958}
Kira Radinsky, Sagie Davidovich, and Shaul Markovitch. 2012.
\newblock \href {https://doi.org/10.1145/2187836.2187958} {Learning causality
  for news events prediction}.
\newblock In \emph{Proceedings of the 21st International Conference on World
  Wide Web}, WWW '12, page 909–918, New York, NY, USA. Association for
  Computing Machinery.

\bibitem[{Radinsky and Horvitz(2013)}]{10.1145/2433396.2433431}
Kira Radinsky and Eric Horvitz. 2013.
\newblock \href {https://doi.org/10.1145/2433396.2433431} {Mining the web to
  predict future events}.
\newblock In \emph{Proceedings of the Sixth ACM International Conference on Web
  Search and Data Mining}, WSDM '13, page 255–264, New York, NY, USA.
  Association for Computing Machinery.

\bibitem[{Reddy et~al.(2019)Reddy, Chen, and Manning}]{reddy-etal-2019-coqa}
Siva Reddy, Danqi Chen, and Christopher~D. Manning. 2019.
\newblock \href {https://doi.org/10.1162/tacl_a_00266} {{C}o{QA}: A
  conversational question answering challenge}.
\newblock \emph{Transactions of the Association for Computational Linguistics},
  7:249--266.

\bibitem[{Rudinger et~al.(2015)Rudinger, Rastogi, Ferraro, and
  Van~Durme}]{rudinger-etal-2015-script}
Rachel Rudinger, Pushpendre Rastogi, Francis Ferraro, and Benjamin Van~Durme.
  2015.
\newblock \href {https://doi.org/10.18653/v1/D15-1195} {Script induction as
  language modeling}.
\newblock In \emph{Proceedings of the 2015 Conference on Empirical Methods in
  Natural Language Processing}, pages 1681--1686, Lisbon, Portugal. Association
  for Computational Linguistics.

\bibitem[{Sandhaus(2008)}]{linguistic2008new}
Evan Sandhaus. 2008.
\newblock \href {https://catalog.ldc.upenn.edu/LDC2008T19} {\emph{The New York
  Times Annotated Corpus}}.
\newblock LDC corpora. Linguistic Data Consortium.

\bibitem[{Srinivasan et~al.(2018)Srinivasan, Arora, and
  Riedl}]{srinivasan-etal-2018-simple}
Siddarth Srinivasan, Richa Arora, and Mark Riedl. 2018.
\newblock \href {https://doi.org/10.18653/v1/N18-2015} {A simple and effective
  approach to the story cloze test}.
\newblock In \emph{Proceedings of the 2018 Conference of the North {A}merican
  Chapter of the Association for Computational Linguistics: Human Language
  Technologies, Volume 2 (Short Papers)}, pages 92--96, New Orleans, Louisiana.
  Association for Computational Linguistics.

\bibitem[{Sun et~al.(2019)Sun, Yu, Yu, and Cardie}]{sun-etal-2019-improving}
Kai Sun, Dian Yu, Dong Yu, and Claire Cardie. 2019.
\newblock \href {https://doi.org/10.18653/v1/N19-1270} {Improving machine
  reading comprehension with general reading strategies}.
\newblock In \emph{Proceedings of the 2019 Conference of the North {A}merican
  Chapter of the Association for Computational Linguistics: Human Language
  Technologies, Volume 1 (Long and Short Papers)}, pages 2633--2643,
  Minneapolis, Minnesota. Association for Computational Linguistics.

\bibitem[{van Dijk(1988)}]{van1988news}
T.A. van Dijk. 1988.
\newblock \href {https://books.google.com/books?id=taRZAAAAMAAJ} {\emph{News as
  Discourse}}.
\newblock Communication (Hillsdale, N.J.). L. Erlbaum Associates.

\bibitem[{Wang et~al.(2020)Wang, Chen, Zhang, and Roth}]{wang-etal-2020-joint}
Haoyu Wang, Muhao Chen, Hongming Zhang, and Dan Roth. 2020.
\newblock \href {https://doi.org/10.18653/v1/2020.emnlp-main.51} {Joint
  constrained learning for event-event relation extraction}.
\newblock In \emph{Proceedings of the 2020 Conference on Empirical Methods in
  Natural Language Processing (EMNLP)}, pages 696--706, Online. Association for
  Computational Linguistics.

\bibitem[{Weber et~al.(2018{\natexlab{a}})Weber, Balasubramanian, and
  Chambers}]{Weber_Balasubramanian_Chambers_2018}
Noah Weber, Niranjan Balasubramanian, and Nathanael Chambers.
  2018{\natexlab{a}}.
\newblock \href {https://ojs.aaai.org/index.php/AAAI/article/view/11932} {Event
  representations with tensor-based compositions}.
\newblock \emph{Proceedings of the AAAI Conference on Artificial Intelligence},
  32(1).

\bibitem[{Weber et~al.(2018{\natexlab{b}})Weber, Shekhar, Balasubramanian, and
  Chambers}]{weber-etal-2018-hierarchical}
Noah Weber, Leena Shekhar, Niranjan Balasubramanian, and Nathanael Chambers.
  2018{\natexlab{b}}.
\newblock \href {https://doi.org/10.18653/v1/D18-1413} {Hierarchical quantized
  representations for script generation}.
\newblock In \emph{Proceedings of the 2018 Conference on Empirical Methods in
  Natural Language Processing}, pages 3783--3792, Brussels, Belgium.
  Association for Computational Linguistics.

\bibitem[{Wilson(2002)}]{wilson2002six}
Margaret Wilson. 2002.
\newblock \href {https://doi.org/10.3758/BF03196322} {Six views of embodied
  cognition}.
\newblock \emph{Psychonomic bulletin \& review}, 9(4):625--636.

\bibitem[{Yao et~al.(2019)Yao, Peng, Weischedel, Knight, Zhao, and
  Yan}]{Yao_Peng_Weischedel_Knight_Zhao_Yan_2019}
Lili Yao, Nanyun Peng, Ralph Weischedel, Kevin Knight, Dongyan Zhao, and Rui
  Yan. 2019.
\newblock \href {https://doi.org/10.1609/aaai.v33i01.33017378} {Plan-and-write:
  Towards better automatic storytelling}.
\newblock \emph{Proceedings of the AAAI Conference on Artificial Intelligence},
  33(01):7378--7385.

\bibitem[{Yarlott et~al.(2018)Yarlott, Cornelio, Gao, and
  Finlayson}]{yarlott-etal-2018-identifying}
W.~Victor Yarlott, Cristina Cornelio, Tian Gao, and Mark Finlayson. 2018.
\newblock \href {https://www.aclweb.org/anthology/W18-4304} {Identifying the
  discourse function of news article paragraphs}.
\newblock In \emph{Proceedings of the Workshop Events and Stories in the News
  2018}, pages 25--33, Santa Fe, New Mexico, U.S.A. Association for
  Computational Linguistics.

\bibitem[{Zhang et~al.(2020)Zhang, Chen, Wang, Song, and
  Roth}]{zhang-etal-2020-analogous}
Hongming Zhang, Muhao Chen, Haoyu Wang, Yangqiu Song, and Dan Roth. 2020.
\newblock \href {https://www.aclweb.org/anthology/2020.emnlp-main.119.pdf}
  {Analogous process structure induction for sub-event sequence prediction}.
\newblock In \emph{Proceedings of the 2020 Conference on Empirical Methods in
  Natural Language Processing (EMNLP)}, pages 1541--1550.

\bibitem[{Zheng et~al.(2020)Zheng, Cai, and Chen}]{10.1145/3397271.3401173}
Jianming Zheng, Fei Cai, and Honghui Chen. 2020.
\newblock \href {https://doi.org/10.1145/3397271.3401173} {Incorporating
  scenario knowledge into a unified fine-tuning architecture for event
  representation}.
\newblock In \emph{Proceedings of the 43rd International ACM SIGIR Conference
  on Research and Development in Information Retrieval}, SIGIR '20, page
  249–258, New York, NY, USA. Association for Computing Machinery.

\bibitem[{Zhou et~al.(2021)Zhou, Richardson, Ning, Khot, Sabharwal, and
  Roth}]{zhou-etal-2021-temporal}
Ben Zhou, Kyle Richardson, Qiang Ning, Tushar Khot, Ashish Sabharwal, and Dan
  Roth. 2021.
\newblock \href {https://doi.org/10.18653/v1/2021.naacl-main.107} {Temporal
  reasoning on implicit events from distant supervision}.
\newblock In \emph{Proceedings of the 2021 Conference of the North American
  Chapter of the Association for Computational Linguistics: Human Language
  Technologies}, pages 1361--1371, Online. Association for Computational
  Linguistics.

\end{thebibliography}
\bibliographystyle{acl_natbib}

\appendix

%\section{Example Appendix}
\label{sec:appendix}

\end{document}